\documentclass{article}
\usepackage{spconf,amsmath,graphicx}
\usepackage{amssymb}
\usepackage{multirow}
\usepackage{hyperref}

\title{Non-Deterministic Face Mask Removal Based On 3D Priors}
\twoauthors
 {Xiangnan Yin, Liming Chen}
	{Ecole Centrale de Lyon, France}
 {Di Huang}
	{Beihang University, China}

\begin{document}
%
\maketitle
\begin{abstract}
This paper presents a novel image inpainting framework for face mask removal. Although current methods have demonstrated their impressive ability in recovering damaged face images, they suffer from two main problems: the dependence on manually labeled missing regions and the deterministic result corresponding to each input. The proposed approach tackles these problems by integrating a multi-task 3D face reconstruction module with a face inpainting module. Given a masked face image, the former predicts a 3DMM-based reconstructed face together with a binary occlusion map, providing dense geometrical and textural priors that greatly facilitate the inpainting task of the latter. By gradually controlling the 3D shape parameters, our method generates high-quality dynamic inpainting results with different expressions and mouth movements. Qualitative and quantitative experiments verify the effectiveness of the proposed method. Our Code: \url{https://github.com/face3d0725/face_de_mask}
\end{abstract}
\section{Introduction}
\label{sec:intro}

Wearing face masks in public has become an essential hygiene practice to control the spread of COVID-19, posing new challenges for face-related computer vision tasks. Computers need to accomplish face recognition, expression recognition, landmark detection, etc., using minimal exposed facial textures. Although many recent studies focus on the masked scenario, most are task-specific and not universally applicable. In comparison, directly restoring mask-occluded face texture promises to be a one-stop solution to the problem. To this end, we need to tackle two sub-tasks: 1) detecting the occluded region and 2) recovering the face textures, corresponding to image segmentation and face image inpainting, respectively.   

Thanks to the revolutionary emergence of deep learning, data-driven approaches have dominated computer vision with great success. However, this also leads to the reliance on high-quality training data. Regarding mask segmentation specifically, large, diverse, and manually annotated mask datasets are in strong demand due to the targets' varying shapes, orientations, and textures. Some methods synthesize training data by overlaying masks on ordinary face images, which is a cheap interim solution before a large paired masked face dataset becomes available. 

\begin{figure*}[!t]
    \centering
    \includegraphics[width=.8\linewidth]{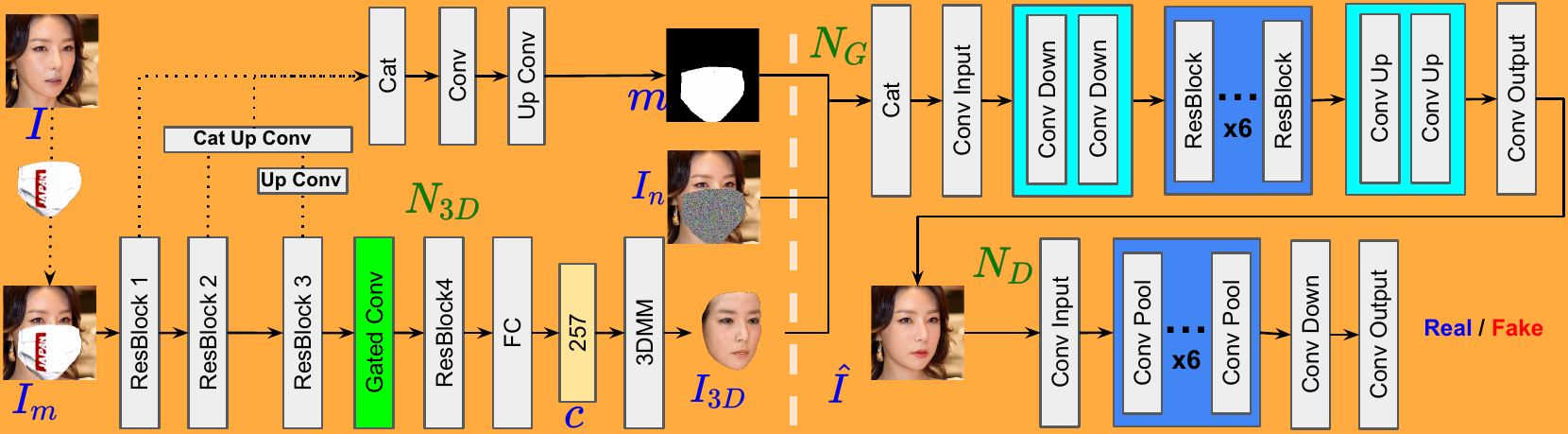}
    \caption{We first synthesize the training data $I_m$ by adding masks to ordinary face images $I$. Next, we use a multi-task model $N_{3D}$ to predict the mask silhouette $m$ and the 3D reconstructed face $I_{3D}$ of the input. Finally, we synthesize the non-occluded face $\hat{I}$ based on the noised face $I_n$, $m$ and $I_{3D}$. A VGG-shaped discriminator $N_D$ is leveraged to distinguish $\hat{I}$ from the real.}
    \label{fig:overview}
\end{figure*}

Early image inpainting methods fill the holes by iteratively searching nearest neighbor textures from the background~\cite{barnes2009patchmatch}. However, such copy-and-paste methods only consider internal information within the image, making them only capable of recovering tiny, smooth textures and not dealing with semantic-level deficiencies such as masked noses and mouths. On the other hand, data-driven approaches learn the data distribution from large datasets, allowing them to restore the semantic-level image patterns. Context Encoder~\cite{pathak2016context} pioneered the adversarial training paradigm. ~\cite{liu2018image,yu2019free} exploits feature masking to deal with free-form missing regions. Also, different attention modules~\cite{zhang2019self,zheng2019pluralistic} have been proposed to break through the limited receptive field of the convolution kernel and thus explicitly model long-distance dependencies. Despite the improved inpainting quality, most methods produce only deterministic results, ignoring multiple fill options.

This paper proposes a novel 3D reconstruction-guided method for removing masks from face images in the wild. The model comprises a multi-task mask-robust 3D face reconstruction module and a face inpainting module. The former predicts both the 3D Morphable Model (3DMM)~\cite{blanz1999morphable} parameters and the binary occlusion map of the masked face, and the latter recovers the missing facial texture conditioned by the rendered 3D prior. By changing the 3DMM parameters, we can control the shape and expression of the recovered face both accurately and smoothly.  

\section{Related works}
\label{sec:related}
The closest work to ours is that of Din \textit{et al.}~\cite{din2020novel}, where we both focus on the problem of face mask removal and divide it into mask segmentation and face painting. Our method surpasses theirs in two aspects: 1) We labeled more mask templates (900 vs. 50) to train the mask segmentation task. 2) Our inpainting results are diverse and highly controllable. 

Some variational autoencoder-based methods can also produce non-deterministic outputs~\cite{sohn2015learning, zheng2019pluralistic} by sampling latent codes from predicted distributions. Although the stochastic nature of the VAE brings about varied results, diversity is never guaranteed: the targets are still fixed, leading to 1) sharp latent distributions and 2) robust decoder to the latent codes' variations, degrading the framework into a common autoencoder. Furthermore, the diversity introduced by random sampling is neither controllable nor smooth. Although some other methods conditioned on sketches~\cite{jo2019sc}, facial landmarks~\cite{yang2019lafin}, or segmentation maps~\cite{yu2020semantic} do yield editable results, the sparsity and instability of such conditions lead to poor controlling accuracy.

\section{Approach}
\label{sec:approach}

Since collecting large amounts of paired with/without the mask face images is infeasible, we train on synthetic data pairs generated by overlaying masks on ordinary face images, as shown on the leftmost of Figure~\ref{fig:overview}. The proposed model is composed of a multi-task 3D face reconstruction-mask segmentation module $N_{3D}$ and a face inpainting module $N_G$, corresponding to the left and right halves of Figure~\ref{fig:overview}, respectively. Given a masked face image $I_m$, $N_{3D}$ predicts its 1) corresponding 3DMM parameters $c$, from which a 3D face $I_{3D}$ could be reconstructed and rendered, and 2) the occlusion mask $m$, indicating the mask silhouette. We then replace the mask texture with random noises based on $m$ and get $I_n$. Finally, $N_G$ predicts the mask-free face $\hat{I}$ conditioned by $I_n$, $m$, and $I_{3D}$. We also employ a discriminator $N_D$ to increase the realism of the generated images. The Following presents each module in detail.  

\subsection{Segmentation-Reconstruction Module}
$N_{3D}$ takes ResNet50~\cite{he2016deep} as its backbone and fulfills 3D face reconstruction and mask segmentation. Intuitively, the neural network captures global shape patterns in the bottom layers and detailed texture patterns in the top layers, while the masks usually occupy a large area of the face with relatively simple textures, so we perform mask segmentation using features from the first three residual blocks. The segmentation task is solely guided by the Binary Cross-Entropy (BCE) loss: 
\begin{equation}
    L_{bce} = -\frac{1}{WH} \sum (m\odot \mathrm{log}(\hat{m})+(1-m)\odot\mathrm{log}(1-\hat{m})),
\end{equation}
where $\hat{m}$ and $m$ denote the predicted and the ground truth binary mask, $W$ and $H$ denote the spatial dimension of the binary map. The ``online hard example mining" (OHEM) technique is also utilized to make the training more efficient. 

To concentrate the model on visible textures while predicting the 3DMM parameters, we also integrate a gated convolution~\cite{yu2019free} layer before the last residual block of ResNet50, predicting a dynamic feature mask for each channel. The 3D reconstruction branch outputs a vector $\hat{c}\in \mathbb{R}^{237}$, containing the face's shape, pose, texture, and illumination parameters. To accelerate the training, we use 3D coefficients predicted from the original unmasked face images by the pre-trained model of ~\cite{deng2019accurate} as the ground truth of $c$. Following losses jointly guide the 3D reconstruction task:  

The most direct term is the coefficient loss. 
\begin{equation}
    L_{coef} = \frac{1}{N}\|\hat{c} - c\|_1,
    \label{eq:2}
\end{equation}
where $\hat{c}$ and $c$ are the predicted and the ground truth 3D coefficients, $N$ denote the dimension of $c$. 

However, the coefficient level loss treats the discrepancy in all dimensions equally, which is unreasonable, as some dimensions affect the reconstruction results much more than others (e.g., poses v.s. illuminations). Hence we introduce the photo loss, which constraints the training at the image level.
\begin{equation}
    L_{photo} = \frac{1}{\sum M} \|I_{3D}\odot M - I \odot M\|_2, 
\end{equation}
where $I_{3D}$ denotes the rendered reconstruction result, $I$ denotes the original face image, and $M$ denotes the binary face region map (provided by the training dataset). 

As with most face reconstruction methods, we apply identity loss for better capturing the face identity. 
\begin{equation}
\label{eq:id}
    L_{id} = 1-\frac{\mathcal{F}(I_{3D})\mathcal{F}(I)}{\|\mathcal{F}(I_{3D})\|\|\mathcal{F}(I)\|},
\end{equation}
where $\mathcal{F}(\cdot)$ denotes the feature extraction operation via a pre-trained Arcface\cite{deng2019arcface} model.

Finally, we leverage landmark loss as~\cite{deng2019accurate} to loosely constrain the shape and pose of the reconstructed face.  
\begin{equation}
    L_{lm} = \frac{1}{n_{pt}}\sum_{i=1}^{n_{pt}}\omega_i\|\hat{q}_i - q_i\|^2,  
\end{equation}
where $\hat{q}_i$ and $q_i$ represent the 68 ($n_{pt}=68$) facial landmarks indexed from the predicted and the ground truth (reconstructed from $c$ in Equation~\ref{eq:2}) 3D faces, respectively. $\omega_i$ is the weight corresponding to the $i$th landmark, set to 20 for the nose and inner-mouth points and 1 for others. 

The overall loss function is formulated as:
\begin{equation}
    L_{3D} = L_{bce} + L_{coef} + L_{photo} + \lambda{id}L_{id} + \lambda_{lm}L_{lm}, 
\end{equation}
where $\lambda_{id}=0.1$ and $\lambda_{lm}=0.001$. 

\subsection{Inpainting Module}

The inpainting module consists of a generator $N_G$ with stacked residual blocks and a discriminator $N_D$ with the VGG structure. As shown in Figure~\ref{fig:overview}, $N_G$ concatenates the mask parsing map $m$, the 3DMM-based face $I_{3D}$, and the noised image $I_{n}$ as input and outputs $\hat{I}$, which recovers the original mask-free face image $I$. Further, $\hat{I}$ and $I$ are fed into $N_D$ to obtain their probabilities of being real data. We utilize the following losses to train the model: 

\noindent \textbf{Pixel-wise loss}, 
\begin{equation}
    L_{pix} = \frac{1}{HWC}\|\hat{I}-I\|_1, 
\end{equation}
where $H$, $W$, $C$ are the height, width and channels of $I$. 

\noindent \textbf{Identity loss} $L_{id}$, formulated the same as Equation~\ref{eq:id}, except replacing $I_{3D}$ therein with $\hat{I}$. 

\noindent \textbf{Total variation loss}~\cite{mahendran2015understanding},
\begin{equation}
    L_{tv} = \frac{1}{HWC} (\|\nabla_x \hat{I}\|^2 +  \|\nabla_y \hat{I}\|^2),
\end{equation}
where $\nabla_\_$ denotes the directional gradient. 

\noindent\textbf{Adversarial loss},
\begin{equation}
    L_{adv} = -\mathbb{E}_{\hat{I}}[\mathrm{log}D(\hat{I}_i)],
\end{equation}
where $D(\cdot)$ denotes the mapping function of $N_D$; the larger its value, the more its input tends to be real. 

\noindent\textbf{The full loss} of $N_G$ is summarized as:
\begin{equation}
    L_G = \lambda_{pix} L_{pix} + \lambda_{id} L_{id} + \lambda_{tv} L_{tv} + \lambda_{adv} L_{adv},
\end{equation}
where $\lambda_{pix}=10$, $\lambda_{id}=0.1$, $\lambda_{tv}=0.1$, $\lambda_{adv}=0.01$. 

\noindent\textbf{Discriminator loss}, the loss of $N_D$ follows the implementation of~\cite{choi2020stargan}, which is composed of an ordinary BCE loss and a zero-centered gradient penalty for real images, 
\begin{equation}
    L_D = \mathbb{E}_I[\mathrm{log}(D(I))] + \mathbb{E}_{\hat{I}}[\mathrm{log}(1-D(\hat{I}))] + \mathbb{E}_I[\nabla_I^2D(I)]
\end{equation}

\section{Experiments}
\label{sec:exp}

We first present our implementation details. Then, we qualitatively compare our method's 3D face reconstruction, mask removal, and face editing abilities with state-of-the-art. Finally, we quantitatively compare the face restoration ability of different methods at pixel and perceptual levels.

\subsection{Training Details}
\begin{table}[t]
    \centering
    \small
    \begin{tabular}{c c c c c c}
    \hline
        \multirow{2}{*}{Methods} & {\footnotesize Hong}  & {\footnotesize ELFW} & {\footnotesize Din} & {\footnotesize Anwar} & \multirow{2}{*}{\footnotesize \textbf{Ours}} \\
        &{\footnotesize \textit{et al.}\cite{hong20213d}} &{\footnotesize ~\cite{redondo2020extended}} & {\footnotesize\textit{et al.}\cite{din2020novel}} & {\footnotesize\textit{et al.}\cite{anwar2020masked}} \\
        \hline 
         Shapes & 14 & 12 & 50 & 20 & \textbf{900} \\
         
         Textures & -- & -- & -- & 27 & \textbf{800}\\
    \hline
    \end{tabular}
    \caption{The mask diversity of different methods or datasets.}
    \label{tab:dataset}
\end{table}

Most mask-related approaches synthesize masked/unmasked training pairs by overlaying mask templates on face images of existing face datasets. However, as Table~\ref{tab:dataset} shows, the mask templates used by previous methods are pretty limited; only a few tens of variations are far from sufficient to train a robust model. Therefore, we 1) manually keyed out 900 masks from the masked face images and 2) collected 800 texture patches to replace the textures of the original masks\footnote{Images are downloaded from Google and labeled using Apple Pencil.}. 

The mask templates are then combined with CelebAMask-HQ~\cite{lee2020maskgan} and FFHQ~\cite{karras2017progressive} to generate data pairs on the fly as training goes (1000 images from FFHQ are left out for testing). We train 500,000 steps for $N_{3D}$ and 200,000 steps for $N_G$-$N_D$, both with a batch size of 8 and an initial learning rate of $1e^{-4}$. For each module, the learning rate drops to  $1e^{-5}$ when the training reaches its midpoint. We use Adam with betas set to $[0.9, 0.999]$ to optimize the two modules. It takes about 60 hours to train $N_{3D}$ and 40 hours to train $N_G$-$N_D$ on two Nvidia GTX 1080 GPUs. 

\begin{figure}
    \centering
    \includegraphics[width=.8\linewidth]{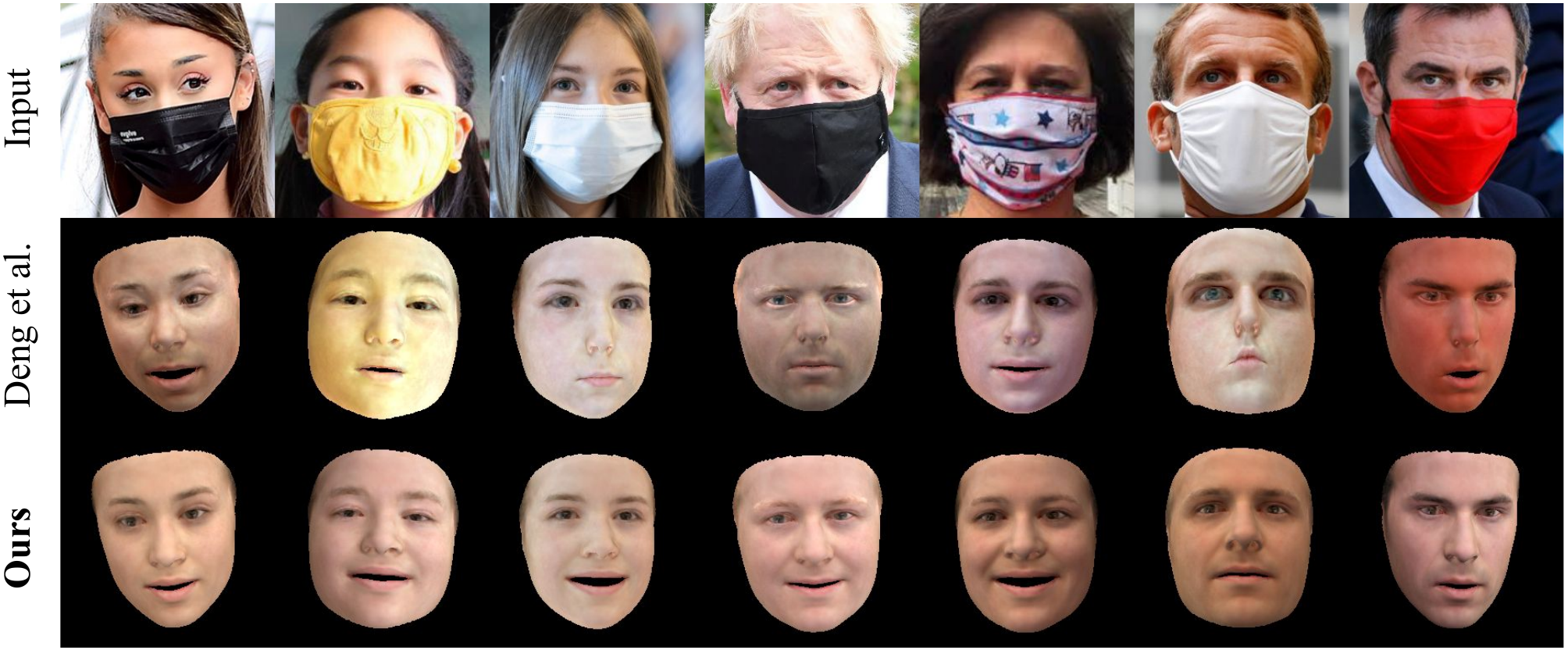}
    \caption{3D face reconstruction ability for masked faces.}
    \label{fig:3d}
\end{figure}

\subsection{Qualitative results}
Accurately reconstructing the 3D face from masked faces is the prerequisite for the success of the subsequent inpainting module. Therefore, we first compare our method with the SOTA 3D reconstruction method of Deng \textit{et al}. As shown in Figure~\ref{fig:3d}, the method of Deng \textit{et al.} is stronly influenced by the mask, resulting in deviations in texture and poses. In comparison, our method is robust to face masks thanks to the synthetic masked face images for training.

\begin{figure}
    \centering
    \includegraphics[width=.8\linewidth]{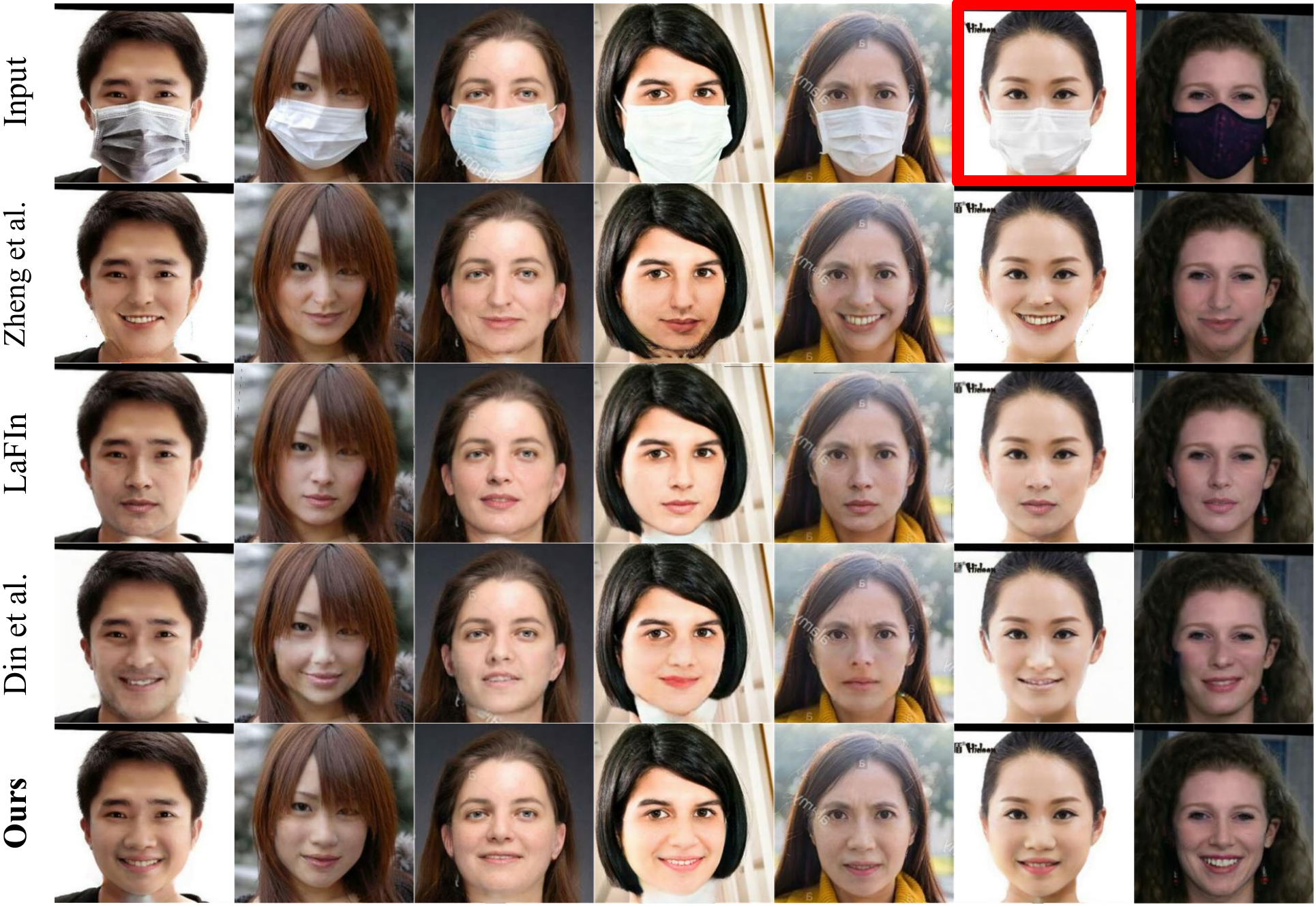}
    \caption{Comparison of mask removal ability, the inputs are aligned according to the methods' settings, and the outputs are remapped to the original images for a consistent view.}
    \label{fig:remove}
\end{figure}

As Section~\ref{sec:related} mentions, the method closest to ours is that of Din \textit{et al.}~\cite{din2020novel}. Unfortunately, they do not release their code; therefore, we use the images from their paper for a more convincing comparison. The other two methods we compare are LaFIn~\cite{yang2019lafin} and Zhang \textit{et al.}, which can generate diverse inpainting results. We provide those methods with mask regions detected by $N_{3D}$. As Figure~\ref{fig:remove} shows, our approach significantly outperforms Zhang \textit{et al.} and Din \textit{et al.}. The random sampling in the hidden space leads to apparent artifacts in the results of Zhang \textit{et al.}. Without a shape prior, the method of Din \textit{et al.} may generate distorted faces (row 4 column 2); In addition, the poor accuracy of their mask segmentation module results in residual mask edges on the face (row 4, columns 6 and 7). Our results are comparable with LaFIn, however, the latter requires additional binary mask maps. 

\begin{figure}
    \centering
    \includegraphics[width=.8\linewidth]{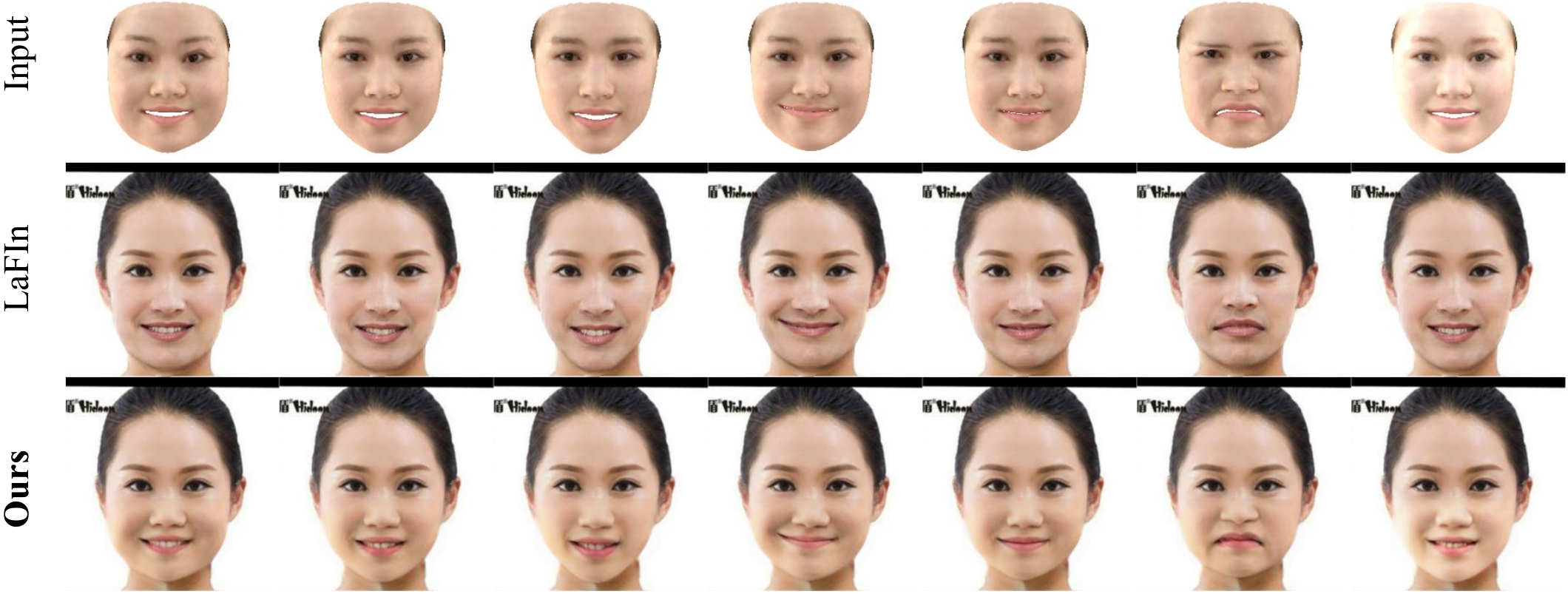}
    \caption{Comparison of face editing ability.}
    \label{fig:edit}
\end{figure}

We further compare the face editing ability of our model with LaFIn, the landmark-guided face inpainting method. This time we provide LaFIn with 68 facial landmarks extracted from our predicted 3D face model. Figure~\ref{fig:edit} shows the results guided by different 3D priors(for the face in the red box in Figure~\ref{fig:remove}). The first six columns are conditioned by different shapes and the last column is conditioned by a brighter skin. As can be seen, with the guidance of our landmarks, LaFIn can generate diverse inpainting results. Nevertheless, due to the sparsity of the landmark, the generated faces do not precisely comply with the 3D face shapes; in addition, LaFIn cannot change the skin color as we do in the last column.  

\subsection{Quantitative results}
\begin{table}[t]
    \centering
    \small
    \begin{tabular}{c c c c c c}
    \hline
         Methods & $L_1 \downarrow$ & PSNR $\uparrow$ & SSIM $\uparrow$ & Cos ID $\uparrow$ & FID $\downarrow$\\
         \hline
         Zheng \textit{et al.} & 0.020 & 24.709 & 0.893 & 0.506 & 12.808 \\
         LaFIn & 0.018 & 25.569 & 0.897 & 0.543 & 12.465 \\
         Din \textit{et al.} & -- & 26.19 & 0.864 & -- & \textbf{3.548} \\
         EdgeConnect$\dagger$ & -- & 20.87 &0.864 & -- & 3.555 \\
         EdgeConnect & 0.018 & 25.305 & 0.895 & 0.537 & 15.27 \\
         \textbf{Ours} & \textbf{0.014} & \textbf{27.230} & \textbf{0.912} & \textbf{0.654} & 9.744 \\
    \hline
    \end{tabular}
    \caption{Quantitative comparison.}
    \label{tab:quant}
\end{table}
We synthesized 1000 masked face images on the test set, and then used LaFIn and Zheng \textit{et al.}'s method to recover the unmasked faces (with externally provided binary mask maps). The face restoration ability is evaluated by: L1 Loss,  PSNR score, SSIM score, FID score\footnote{Using the open source tool: https://github.com/mseitzer/pytorch-fid}, and the cosine similarity of the identity features extracted by~\cite{deng2019arcface}. For Din \textit{et al.}'s method, since their code is not publicly available, we adopt the data from their paper. Results are shown in Table~\ref{tab:quant}. As can be seen, our method outperforms others in all metrics except FID score where Din \textit{et al.} reported a much better result. However, Din \textit{et al.}'s data are questionable. To demonstrate this, we further tested EdgeConnect~\cite{Nazeri_2019_ICCV}, a method Din \textit{et al.} compared in their paper, using the released the code and the pre-trained models. The results reported by Din \textit{et al.} and ours are shown in the 4th and 5th rows of Table~\ref{tab:quant}, respectively. It can be seen that the FID score of EdgeConnect reported by Din \textit{et al.} is similar to their proposed method but much lower than the one we tested. We, therefore, question the credibility of Din \textit{et al.}'s data. 

\section{Conclusion}
This paper proposes a novel framework for removing the mask from the face image. First, we manually labeled a large, high-quality dataset of face masks for synthesizing training pairs. Next, we trained a mask-robust multi-task module for reconstructing 3D faces and detecting the mask region of face images. Finally, we proposed a 3D reconstruction guided face inpainting module to generate non-deterministic and highly-controllable results. The proposed method outperforms the state-of-the-art qualitatively and quantitatively.

\bibliographystyle{IEEEbib}
\bibliography{strings,refs}

\end{document}